\documentclass[11pt,a4paper]{article}
\usepackage[hyperref]{acl2018}
\usepackage{times}
\usepackage{latexsym}
\usepackage{url}
\usepackage[utf8]{inputenc} % allow utf-8 input
\usepackage[T1]{fontenc}    % use 8-bit T1 fonts
\usepackage{amsfonts}       % blackboard math symbols
\usepackage{microtype}      % microtypography
\usepackage{color}
\usepackage{graphicx,booktabs,multirow,latexsym,amsmath,bm}

\usepackage{subcaption}
\usepackage{qtree}

\aclfinalcopy % Uncomment this line for the final submission
\def\aclpaperid{1} %  Enter the acl Paper ID here

\title{Latent Tree Learning with Differentiable Parsers:\\Shift-Reduce Parsing and Chart Parsing}

\author{Jean Maillard, Stephen Clark \\
Computer Laboratory, University of Cambridge\\
\texttt{jean@maillard.it, sc609@cam.ac.uk}}

\date{}

\begin{document}
\maketitle
\begin{abstract}
Latent tree learning models represent sentences by composing their words according to an induced parse tree, all based on a downstream task. These models often outperform baselines which use (externally provided) syntax trees to drive the composition order. This work contributes (a) a new latent tree learning model based on shift-reduce parsing, with competitive downstream performance and non-trivial induced trees, and (b) an analysis of the trees learned by our shift-reduce model and by a chart-based model.
\end{abstract}

\section{Introduction}

Popular recurrent neural networks in NLP, such as the Gated Recurrent Unit \citep{gru} and Long Short-Term Memory \citep{lstm}, compute sentence representations by reading their words in a sequence. In contrast, the Tree-LSTM architecture \citep{tree_lstm} processes words according to an input parse tree, and manages to achieve improved performance on a number of linguistic tasks.

Recently, \citet{yogatama_rl_16}, \citet{maillard}, and \citet{choi} all proposed sentence embedding models which work similarly to a Tree-LSTM, but do not require any parse trees as input. These models function without the assistance of an external automatic parser, and without ever being given any syntactic information as supervision. Rather, they induce parse trees by training on a downstream task such as natural language inference. At the heart of these models is a mechanism to assign trees to sentences -- effectively, a natural language parser.
\citet{williams} have recently investigated the tree structures induced by two of these models, trained for a natural language inference task. Their analysis showed that \citet{yogatama_rl_16} learns mostly trivial left-branching trees, and has inconsistent performance; while \citet{choi} outperforms all baselines (including those using trees from conventional parsers), but learns trees that do not correspond to those of conventional treebanks.

In this paper, we propose a new latent tree learning model. Similarly to \citet{yogatama_rl_16}, we base our approach on shift-reduce parsing. Unlike their work, our model is trained via standard backpropagation, which is made possible by exploiting beam search to obtain an approximate gradient. We show that this model performs well compared to baselines, and induces trees that are not as trivial as those learned by the Yogatama et al. model in the experiments of \citet{williams}.

This paper also presents an analysis of the trees learned by our model, in the style of \citet{williams}. We further analyse the trees learned by the model of \citet{maillard}, which had not yet been done, and perform evaluations on both the SNLI data \citep{snli} and the MultiNLI data \citep{mnli}. The former corpus had not been used for the evaluation of trees of \citet{williams}, and we find that it leads to more consistent induced trees.

\section{Related work}
\label{sec:related_work}

The first neural model which learns to both parse a sentence and embed it for a downstream task is by \citet{rae}. The authors train the model's parsing component on an auxiliary task, based on recursive autoencoders, while the rest of the model is trained for sentiment analysis.

\citet{bowman_spinn_16} propose the ``Shift-reduce Parser-Interpreter Neural Network'', a model which obtains syntax trees using an integrated shift-reduce parser (trained on gold-standard trees), and uses the resulting structure to drive composition with Tree-LSTMs.

\citet{yogatama_rl_16} is the first model to jointly train its parsing and sentence embedding components. They base their model on shift-reduce parsing. Their parser is not differentiable, so they rely on reinforcement learning for training.

\citet{maillard} propose an alternative approach, inspired by CKY parsing. The algorithm is made differentiable by using a soft-gating approach, which approximates discrete candidate selection by a probabilistic mixture of the constituents available in a given cell of the chart. This makes it possible to train with backpropagation.

\citet{choi} use an approach similar to easy-first parsing. The parsing decisions are discrete, but the authors use the Straight-Through Gumbel-Softmax estimator \citep{gumbel} to obtain an approximate gradient and are thus able to train with backpropagation.

\citet{williams} investigate the trees produced by \citet{yogatama_rl_16} and \citet{choi} when trained on two natural language inference corpora, and analyse the results. They find that the former model induces almost entirely left-branching trees, while the latter performs well but has inconsistent trees across re-runs with different parameter initializations. 

A number of other neural models have also been proposed which create a tree encoding during parsing, but unlike the above architectures rely on traditional parse trees. \citet{le_forest_15} propose a sentence embedding model based on CKY, taking as input a parse forest from an automatic parser. \citet{dyer_rnng_16} propose RNNG, a probabilistic model of phrase-structure trees and sentences, with an integrated parser that is trained on gold standard trees.

\section{Models}
\label{sec:models}

\paragraph{CKY} The model of \citet{maillard} is based on chart parsing, and effectively works like a CKY parser \cite{C,K,Y} using a grammar with a single nonterminal $A$ with rules $A \to A~A$ and $A \to \alpha$, where $\alpha$ is any terminal. The parse chart is built bottom-up incrementally, like in a standard CKY parser. When ambiguity arises, due to the multiple ways to form a constituent, all options are computed using a Tree-LSTM, and scored. The constituent is then represented as a weighted sum of all possible options, using the normalised scores as weights. In order for this weighted sum to approximate a discrete selection, a temperature hyperparameter is used in the softmax. This process is repeated for the whole chart, and the sentence representation is given by the topmost cell.

We noticed in our experiments that the weighted sum still occasionally assigned non-trivial weight to more than one option. The model was thus able to utilize multiple inferred trees, rather than a single one, which would have potentially given it an advantage over other latent tree models. Hence for fairness, in our experiments we replace the softmax-with-temperature of \citet{maillard} with a softmax followed by a straight-through estimator \citep{ste}. In the forward pass, this approach is equivalent to an argmax function; while in the backward pass it is equivalent to a softmax. Effectively, this means that a single tree is selected during forward evaluation, but the training signal can still propagate to every path during backpropagation. This change did not noticeably affect performance on development data.

\paragraph{Beam Search Shift-Reduce} We propose a model based on beam search shift-reduce parsing (BSSR). The parser works with a queue, which holds the embeddings for the nodes representing individual words which are still to be processed; and a stack, which holds the embeddings of the nodes which have already been computed.  A standard binary Tree-LSTM function \citep{tree_lstm} is used to compute the $d$-dimensional embeddings of nodes:
\noindent
\begin{align*}
\begin{bmatrix}\textstyle\bm{i}\\ \textstyle\bm{f}_L\\ \textstyle\bm{f}_R\\ \textstyle\bm{u}\\ \textstyle\bm{o}\end{bmatrix} &= \mathbf{W}\bm{w} + \mathbf{U} \bm{h}_L + \mathbf{V}\bm{h}_R + \bm{b},\\
\bm{c}\quad &= \bm{c}_L \odot \sigma (\bm{f}_L) + \bm{c}_R \odot \sigma (\bm{f}_R)\\
&\qquad+ \tanh (\bm{u}) \odot \sigma (\bm{i}),\\
\bm{h}\quad &= \sigma(\bm{o}) \odot \tanh ( \bm{c}),
\end{align*}
where $\mathbf{W},\mathbf{U}$ are learned $5d\times d$ matrices, and $\bm{b}$ is a learned $5d$ vector. The $d$-dimensional vectors $\sigma(\bm{i}), \sigma(\bm{f}_L), \sigma(\bm{f}_R)$ are known as \emph{input} gate and \emph{left-} and \emph{right-forget} gates, respectively. $ \sigma(\bm{o}_t)$ and $\tanh(\bm{u}_t)$ are known as \emph{output} gate and \emph{candidate update}. The vector $\bm{w}$ is a word embedding, while $\boldsymbol{h}_L,\boldsymbol{h}_R$ and $\boldsymbol{c}_L,\boldsymbol{c}_R$ are the childrens' $\boldsymbol{h}$- and $\boldsymbol{c}$-states. At the beginning, the queue contains embeddings for the nodes corresponding to single words. These are obtained by computing the Tree-LSTM with $\bm{w}$ set to the word embedding, and $\bm{h}_{L/R},\bm{c}_{L/R}$ set to zero. When a \textsc{shift} action is performed, the topmost element of the queue is popped, and pushed onto the stack. When a \textsc{reduce} action is performed, the top two elements of the stack are popped. A new node is then computed as their parent, by passing the children through the Tree-LSTM, with $\bm{w}=0$. The new node is then pushed onto the stack.

Parsing actions are scored with a simple multi-layer perceptron, which looks at the top two stack elements and the top queue element:
\begin{align*}
\bm{r} &= \mathbf{W}_{s1}\cdot\bm{h}_{s1} + \mathbf{W}_{s2}\cdot\bm{h}_{s2} + \mathbf{W}_{q}\cdot\bm{h}_{q1},\\
\bm{p} &= \operatorname{softmax}\,(\bm{a} + \mathbf{A} \cdot \tanh \bm{r}),
\end{align*}
where $\bm{h}_{s1},\bm{h}_{s2},\bm{h}_{q1}$ are the $\bm{h}$-states of the top two elements of the stack and the top element of the queue, respectively. The three $\mathbf{W}$ matrices have dimensions $d\times d$ and are learned; $\bm{a}$ is a learned 2-dimensional vector; and $\mathbf{A}$ is a learned $2\times d$ vector. The final scores are given by $\log \bm{p}$, and the best action is greedily selected at every time step. The sentence representation is given by the $\bm{h}$-state of the top element of the stack after $2n-1$ steps.

In order to make this model trainable with gradient descent, we use beam search to select the $b$ best action sequences, where the score of a sequence of actions is given by the sum of the scores of the individual actions. The final sentence representation is then a weighted sum of the sentence representations from the elements of the beam. The weights are given by the respective scores of the action sequences, normalised by a softmax and passed through a straight-through estimator. This is equivalent to having an argmax on the forward pass, which discretely selects the top-scoring beam element, and a softmax in the backward pass. 

\begin{table}[t!]
\small
\centering
\begin{tabular}{lrrr}  
\toprule
\textbf{Model} & \textbf{SNLI} & \textbf{MultiNLI+} \\
\midrule
\multicolumn{3}{c}{Prior work: Baselines} \\
\midrule
100D LSTM (Yogatama) & 80.2 & ---\\
300D LSTM (Williams) & 82.6 & 69.1 \\
100D Tree-LSTM (Yogatama) & 78.5 & --- \\
300D SPINN (Williams) & 82.2 & 67.5 \\
\midrule
\multicolumn{3}{c}{Prior work: Latent Tree Models} \\
\midrule
100D ST-Gumbel (Choi) & 81.9 & --- \\
300D ST-Gumbel (Williams) & 83.3 & \textbf{69.5} \\
300D ST-Gumbel$^{\dagger}$ (Williams) & \textbf{83.7} & 67.5 \\
100D CKY (Maillard) & 81.6 & --- \\
100D RL-SPINN (Yogatama) & 80.5 & --- \\
300D RL-SPINN$^{\dagger}$ (Williams) & 82.3 & 67.4 \\
\midrule
\multicolumn{3}{c}{This work: Latent Tree Models} \\
\midrule
100D CKY (Ours) & 82.2 & 69.1 \\
100D BSSR (Ours) & 83.0 & 69.0 \\
\bottomrule
\end{tabular}
\caption{SNLI and MultiNLI (matched) test set accuracy. $\dagger$: results are for the model variant without the leaf RNN transformation.}
\label{tbl:acc}

\end{table}

\begin{table*}[t!]
\small
\centering
\begin{tabular}{clcrrrrrr}  
\toprule
& & & \multicolumn{6}{c}{\textbf{F1 w.r.t.}} \\
& & & \multicolumn{2}{c}{\textbf{Left Branching}} & \multicolumn{2}{c}{\textbf{Right Branching}} & \multicolumn{2}{c}{\textbf{Stanford Parser}} \\
\textbf{Dataset} &\textbf{Model} & \textbf{Self-F1} & \multicolumn{1}{c}{$\bm{\mu}$ ($\bm{\sigma}$)} & \textbf{max} & \multicolumn{1}{c}{$\bm{\mu}$ ($\bm{\sigma}$)} & \textbf{max} & \multicolumn{1}{c}{$\bm{\mu}$ ($\bm{\sigma}$)} & \textbf{max}\\
\midrule
MultiNLI+ & 300D SPINN (Williams) & 71.5 & 19.3 (0.4) & 19.8 & 36.9 (3.4) & \textbf{42.6} & \textbf{70.2} (3.6) & \textbf{74.5}  \\
MultiNLI+ & 300D ST-Gumbel (Williams) & 49.9 & 32.6 (2.0) & 35.6 & \textbf{37.5} (2.4) & {40.3} & 23.7 (0.9) & 25.2 \\
MultiNLI+ & 300D ST-Gumbel$^\dagger$ (Williams) & 41.2 & 30.8 (1.2) & 32.3 & 35.6 (3.3) & 39.9 & 27.5 (1.0) & 29.0 \\
MultiNLI+ & 300D RL-SPINN$^\dagger$ (Williams) & \textbf{98.5} & \textbf{99.1} (0.6) & \textbf{99.8} & 10.7 (0.2) & 11.1 & 18.1 (0.1) & 18.2 \\
MultiNLI+ & 100D CKY (Ours) & 45.9 & 32.9 (1.9) & 35.1 & 31.5 (2.3) & 35.1 & 23.7 (1.1) & 25.0\\
MultiNLI+ & 100D BSSR (Ours) & 46.6 & 40.6 (6.5) & 47.6 & 24.2 (6.0) & 27.7 & 23.5 (1.8) & 26.2\\
MultiNLI+ & \emph{Random Trees} (Williams) & 32.6 & 27.9 (0.1) & 27.9 & 28.0 (0.1) & 28.1 & 27.0 (0.1) & 27.1\\
\midrule
SNLI & 100D RL-SPINN (Yogatama) & --- & \multicolumn{1}{c}{---} & 41.4 & \multicolumn{1}{c}{---} & 19.9 & \multicolumn{1}{c}{---} & \textbf{41.7} \\
SNLI & 100D CKY (Ours) & 59.2 & 43.9 (2.2) & 46.9 & \textbf{33.7} (2.6) & \textbf{36.7} & 30.3 (1.1) & 32.1 \\
SNLI & 100D BSSR (Ours) & \textbf{60.0} & \textbf{48.8} (5.2) & \textbf{53.9} & 26.5 (6.9) & 34.0 & \textbf{32.8} (3.5) & 36.4\\
SNLI & \emph{Random Trees} (Ours) & 35.9 & 32.3 (0.1) & 32.4 & 32.5 (0.1) & 32.6 &  32.3 (0.1) & 32.5\\
\bottomrule
\end{tabular}
\caption{Unlabelled F1 scores of the trees induced by various models against: other runs of the same model, fully left- and right-branching trees, and Stanford Parser trees provided with the datasets. The baseline results on MultiNLI are from \citet{williams}. $\dagger$: results are for the model variant without the leaf RNN transformation.}
\label{tbl:f1}

\end{table*}

\section{Experimental Setup}
\label{sec:setup}

\paragraph{Data}  To  match the settings of \citet{maillard}, we run experiments with the SNLI corpus \citep{snli}. We additionally run a second set of experiments with the MultiNLI data \citep{mnli}, and to match \citet{williams} we augment the MultiNLI training data with the SNLI training data. We call this augmented training set \emph{MultiNLI+}. For the MultiNLI+ experiments, we use the \emph{matched} versions of the development and test sets. We use pre-trained 100D GloVe embeddings\footnote{\scriptsize \url{https://nlp.stanford.edu/projects/glove/}} \citep{glove} for performance reasons, and fine-tune them during training. Unlike \citet{williams}, we do not use a bidirectional leaf transformation. Models are optimised with Adam \citep{adam}, and we train five instances of every model. For BSSR, we use a beam size of 50, and let it linearly decrease to its final size of 5 over the first two epochs.

\paragraph{Setup}  To assign the labels of \emph{entails}, \emph{contradicts}, or \emph{neutral} to the pairs of sentences, we follow \citet{yogatama_rl_16} and concatenate the two sentence embeddings, their element-wise product, and their squared Euclidean distance into a vector $\bm{v}$. We then calculate $\bm{q} = \operatorname{ReLU}\,(\mathbf{C}\cdot\bm{v}+\bm{c})$, where $\mathbf{C}$ is a $200\times 4d$ learned matrix and $\bm{c}$ a 200-dimensional learned bias; and finally predict $p(y=c\mid\bm{q}) \propto \operatorname{exp}\,(\mathbf{B}\cdot\bm{q}+\bm{b})$ where $\mathbf{B}$ is a $3\times 200$ matrix and $\bm{b}$ is 3-dimensional.

\section{Experiments}
\label{sec:experiments}

For each model and dataset, we train five instances using different random initialisations, for a total of $2\times2\times5=20$ instances.

\paragraph{NLI Accuracy} We measure SNLI and MultiNLI test set accuracy for CKY and BSSR. The aim is to ensure that they perform reasonably, and are in line with other latent tree learning models of a similar size and complexity. Results for the best models, chosen based on development set performance, are reported in Table \ref{tbl:acc}.

While our models do not reach the state of the art, they perform at least as well as other latent tree models using 100D embeddings, and are competitive with some 300D models. They also outperform the 100D Tree-LSTM of \citet{yogatama_rl_16}, which is given syntax trees, and match or outperform 300D SPINN, which is explicitly trained to parse.

\paragraph{Self-consistency} Next, we examine the consistency of the trees produced for the development sets. Adapting the code of \citet{williams}, we measure the models' \emph{self F1}, defined as the unlabelled F1 between trees by two instances of the same model (given by different random initializations), averaged over all possible pairs. Results are shown in Table \ref{tbl:f1}. In order to test whether BSSR and CKY learn similar grammars, we calculate the \emph{inter-model F1}, defined as the unlabelled F1 between instances of BSSR and CKY trained on the same data, averaged over all possible pairs. We find an average F1 of 42.6 for MultiNLI+ and 55.0 for SNLI, both above the random baseline.

Our Self F1 results are all above the baseline of random trees. For MultiNLI+, they are in line with ST-Gumbel. Remarkably, the models trained on SNLI are noticeably more self-consistent. This shows that the specifics of the training data play an important role, even when the downstream task is the same. A possible explanation is that MultiNLI has longer sentences, as well as multiple genres, including telephone conversations which often do not constitute full sentences \citep{mnli}. This would require the models to learn how to parse a wide variety of styles of data.
It is also interesting to note that the inter-model F1 scores are not much lower than the self F1 scores. This shows that, given the same training data, the grammars learned by the two different models are not much more different than the grammars learned by two instances of the same model.

\paragraph{F1 Scores} Finally, we investigate whether these models learn grammars that are recognisably left-branching, right-branching, or similar to the trees produced by the Stanford Parser which are included in both datasets. We report the unlabelled F1 between these and the trees from from our models in Table \ref{tbl:f1}, averaged over the five model instances. We show mean, standard deviation, and maximum.

We find a slight preference from BSSR and the SNLI-trained CYK towards left-branching structures. Our models do not learn anything that resembles the trees from the Stanford Parser, and have an F1 score with them which is at or below the random baseline. Our results match those of \citet{williams}, which show that whatever these models learn, it does not resemble PTB grammar.

\section{Conclusions}

First, we proposed a new latent tree learning model based on a shift-reduce parser. Unlike a previous model based on the same parsing technique, we showed that our approach does not learn trivial trees, and performs competitively on the downstream task.

Second, we analysed the trees induced by our shift-reduce model and a latent tree model based on chart parsing. Our results confirmed those of previous work on different models, showing that the learned grammars do not resemble PTB-style trees \citep{williams}. Remarkably, we saw that the two different models tend to learn grammars which are not much more different than those learned by two instances of the same model.

Finally, our experiments highlight the importance of the choice of training data used for latent tree learning models, even when the downstream task is the same. Our results suggest that MultiNLI, which has on average longer sentences coming from different genres, might be hindering the current models' ability to learn consistent grammars. For future work investigating this phenomenon, it may be interesting to train models using only the written genres parts of MultiNLI, or MultiNLI without the SNLI corpus.

\section*{Acknowledgments}

We are grateful to Chris Dyer for the several productive discussions. We would like to thank the anonymous reviewers for their helpful comments.

\bibliographystyle{acl_natbib}
\bibliography{main}

\end{document}